\pdfoutput=1

\documentclass[11pt]{article}

\usepackage[final]{acl}

\usepackage{times}
\usepackage{latexsym}

\usepackage[T1]{fontenc}

\usepackage[utf8]{inputenc}

\usepackage{microtype}

\usepackage{inconsolata}

\usepackage{graphicx}
\usepackage{xspace}
\usepackage{caption}
\usepackage{titlesec}
\titlespacing{\subsection}{0pt}{4pt plus 1pt minus 1pt}{2pt plus 1pt minus 1pt}
\usepackage{makecell}

\usepackage{multirow}
\usepackage{amsmath, amssymb}
\usepackage{subcaption}

\definecolor{darkpastelgreen}{rgb}{0.01, 0.75, 0.24}
\definecolor{blue}{rgb}{0.0, 0.0, 1.0}

\definecolor{mycolor}{rgb}{0.0, 0.0, 0.0}

\usepackage{xspace}
\newcommand{\method}[0]{\textsc{GENUINE}\xspace}

\title{GENUINE: Graph Enhanced Multi-level Uncertainty Estimation for Large Language Models}

\author{
 \textbf{Tuo Wang\textsuperscript{1}},
 \textbf{Adithya Kulkarni\textsuperscript{2}},
 \textbf{Tyler Cody\textsuperscript{1}},
 \textbf{Peter A. Beling\textsuperscript{1}},
 \textbf{Yujun Yan\textsuperscript{3}},
 \textbf{Dawei Zhou\textsuperscript{1}}
\\
\\
 \textsuperscript{1}Virginia Polytechnic Institute and State University,
 \textsuperscript{2} Ball State University,
 \textsuperscript{3}Dartmouth College
\\
 \small{
   \textbf{Correspondence:} \href{mailto:tuowang@vt.edu}{tuowang@vt.edu}
 }
}

\begin{document}
\maketitle
\begin{abstract}
    Uncertainty estimation is essential for enhancing the reliability of Large Language Models (LLMs), particularly in high-stakes applications. Existing methods often overlook semantic dependencies, relying on token-level probability measures that fail to capture structural relationships within the generated text. We propose GENUINE: \textbf{G}raph \textbf{EN}hanced m\textbf{U}lti-level uncerta\textbf{IN}ty \textbf{E}stimation for Large Language Models, a structure-aware framework that leverages dependency parse trees and hierarchical graph pooling to refine uncertainty quantification. By incorporating supervised learning, \method effectively models semantic and structural relationships, improving confidence assessments. Extensive experiments across NLP tasks show that \method achieves up to 29\% higher AUROC than semantic entropy-based approaches and reduces calibration errors by over 15\%, demonstrating the effectiveness of graph-based uncertainty modeling. The code is available at \url{https://github.com/ODYSSEYWT/GUQ}.
\end{abstract}
\section{Introduction}

Large Language Models (LLMs) have demonstrated remarkable capabilities in conversation~\cite{wu2023autogenenablingnextgenllm}, logical reasoning~\cite{wang2023chatgptdefendbelieftruth}, and scientific discovery~\cite{shojaee2024llm}. Models such as GPT-4~\cite{achiam2023gpt}, Gemini~\cite{team2023gemini}, and DeepSeek~\cite{liu2024deepseek}, trained on vast corpora and aligned to human preferences, have significantly expanded the potential of AI. However, despite these advancements, LLMs are prone to well-documented reliability issues, including hallucinations and factual inaccuracies~\cite{10.1145/3703155, liu2024trustworthyllmssurveyguideline}. These issues pose serious risks, particularly in high-stakes applications such as medical diagnosis~\cite{panagoulias2024evaluating}, financial decision-making~\cite{de2023optimized}, and legal advisory systems~\cite{cheong2024not}, where users must rely on the model's outputs with confidence. Therefore, uncertainty quantification (UQ), which assesses the trustworthiness of an LLM response, is essential for safe and effective human and artificial intelligence interaction.
\begin{figure}[t!]
    \centering
    \includegraphics[width=\linewidth]{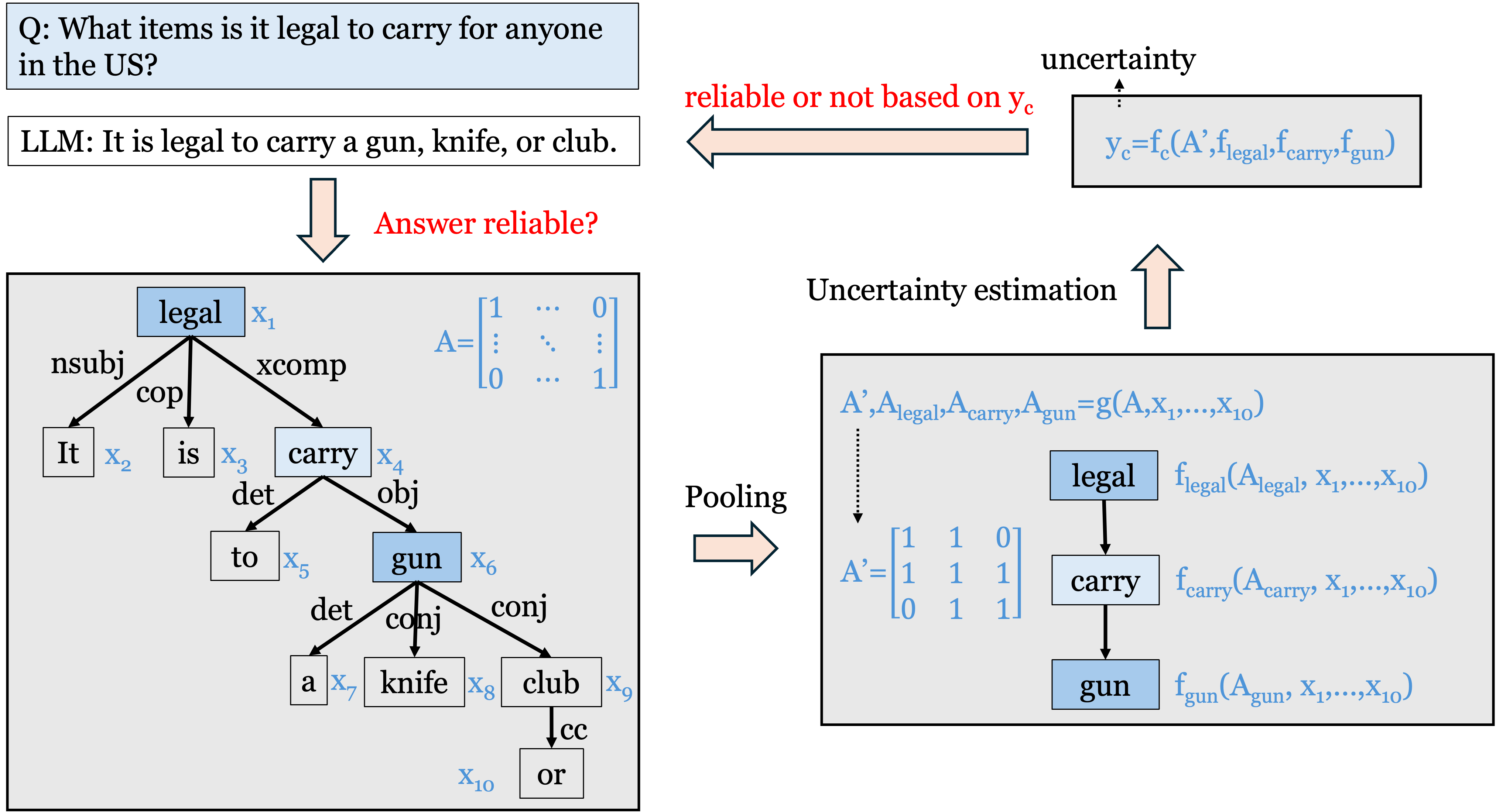}
    \caption{An example highlighting the role of graph pooling to identify tokens' significance in uncertainty estimation. Critical tokens are identified by graph pooling through dependency parsing tree and backpropagation of ground truth label, which makes the uncertainty task aware. A represents the adjacency matrix for a tree structure, where connected tree nodes are assigned value 1 while others are assigned value 0. g represents the pooling method, and f represents the information propagating through the pooling process.}
    \label{fig: motivation example}
\end{figure}

UQ in LLM-generated outputs presents several challenges. First, LLMs often produce long-form textual responses, making attributing uncertainty to specific components difficult. Second, uncertainties may affect only a few critical tokens within an otherwise coherent response, undermining the reliability of the entire output. Third, aggregating uncertainty across multiple tokens in lengthy outputs is non-trivial, requiring distinguishing semantically pivotal tokens from those not pivotal.

Previous studies have explored various approaches to quantify uncertainty in LLM outputs. Some methods rely on self-evaluation through modified prompts~\cite{tian2023just}, though they often inherit the model's biases. Others use token-level uncertainty measures based on logits, entropy, or probability distributions~\cite{kuhn2023semantic, malinin2020uncertainty, malinin2021uncertaintyestimationautoregressivestructured}. Recent advancements, such as semantic entropy, cluster semantically equivalent generations and measure entropy as an uncertainty indicator~\cite{kuhn2023semantic}. However, most existing methods treat all tokens equally, overlooking findings that certain tokens carry more semantic weight in determining output validity~\cite{liu2024uncertainty, duan2024shifting, cheng-vlachos-2024-measuring}. Recently, some approaches~\cite{duan2024shifting} emphasize the different contributions of each token in uncertainty estimation. \cite{duan2024shifting} depends on external, smaller models to estimate token importance. While admitting that these smaller models can capture the semantics between different tokens. They inherently operate independently of the LLM's internal representations, which introduce inconsistencies, misinterpret token dependencies~\cite{zhang2025bert}. 
Besides, using an external model to estimate semantic importance also adds significantly more computational cost and latency, especially in longer contexts. With the recognition of these limitations, we propose to use semantic parsing trees to help identify the contributions of different tokens, as they purely represent the semantic meaning between different tokens, and recent studies~\cite{jin2024analyzing, an-etal-2025-rethinking} have demonstrated that semantic structure can further improve LLMs' performance.

To illustrate this issue, consider the example in Fig.~\ref{fig: motivation example}. A user inquires about legal items to carry in the United States, but the model responds with a list of illegal items, such as a gun, knife, and club. The misunderstanding stems from the token "legal," which is central to the query's meaning. A minor modification, replacing the word legal with illegal, would render the response appropriate. This example underscores two insights: certain tokens are disproportionately influential in determining output validity. Dependency parse trees effectively capture the hierarchical structure of sentence meaning by identifying core decision points. Building on these insights, we propose leveraging dependency parse trees and graph pooling techniques to infer LLM prediction uncertainty in a structured and interpretable manner.

Modeling uncertainty estimation as a graph-based problem offers several advantages. Graphs inherently capture dependencies between generated tokens, reflecting the autoregressive nature of LLMs, where each token influences subsequent ones. By representing an LLM response as a structured graph, we can propagate and aggregate critical information across tokens, ensuring that semantically significant tokens contribute more substantially to the overall uncertainty estimate. However, this approach introduces several challenges. Determining the optimal graph structure that accurately represents token dependencies remains an open question. Selecting appropriate graph pooling techniques that summarize uncertainty information effectively without losing essential context is difficult. Addressing these challenges is essential to fully realize the potential of graph-based uncertainty estimation.

Our approach integrates multiple uncertainty features to enhance robustness. Specifically, we utilize probability distributions, entropy-based measures, and LLM embeddings to model uncertainty. We introduce a hierarchical strategy to address the challenge of aggregating uncertainty over long-form text. We construct a dependency parse tree for each sentence to extract structural and semantic relationships. We merge sentence-level trees into a document-level graph by connecting their root nodes. We apply graph pooling techniques to model uncertainty across the entire paragraph efficiently. \method involves learning pooling functions that adaptively fuse different features, capturing both local and global dependencies within the text. Experimental results prove that \method outperforms other baselines, highlighting the critical role of structural relationships in uncertainty estimation. Furthermore, we compare the effectiveness of probability-based and embedding-based features across various datasets and LLMs, offering insights into their respective utilities. Given that commercial LLMs typically provide only probability-related features, our findings suggest an intriguing direction for future research. Exploring whether open-source LLMs, which offer both probability and embedding features, can facilitate superior UQ compared to their commercial counterparts.

The following are our main contributions: \\
\noindent $\bullet$ We highlight the role of semantically significant tokens in uncertainty estimation, demonstrating how structural relationships can enhance model uncertainty assessment.\\
\noindent $\bullet$ We propose a graph-based framework for LLM UQ, integrating dependency parse trees and graph pooling to capture structural and semantic relationships in the generated text.\\
\noindent $\bullet$ We develop an adaptive graph pooling mechanism that effectively propagates and aggregates uncertainty information by learning to fuse multiple uncertainty features.\\
\noindent $\bullet$ We conduct extensive experiments on real-world datasets, demonstrating that \method outperforms existing UQ methods in assessing the trustworthiness of LLM-generated responses.
\section{Related Works}
\noindent \textbf{Uncertainty Quantification in LLMs.} \textcolor{mycolor}{Uncertainty quantification is well-studied in traditional machine learning~\cite{chen2019embedding,zhao2020uncertainty}, but remains challenging for LLMs due to their open-ended outputs, where multiple valid responses can exist. This flexibility complicates uncertainty estimation, requiring methods beyond standard predictive confidence.
Current approaches fall into two categories. Self-assessment prompts LLMs to estimate their own uncertainty~\cite{kadavath2022languagemodelsmostlyknow, lin2022teachingmodelsexpressuncertainty, tian-etal-2023-just}, but often reflect model biases and inconsistencies. External methods assess uncertainty via output consistency~\cite{manakul-etal-2023-selfcheckgpt} or entropy measures~\cite{malinin2020uncertainty}, though these typically assume uniform token importance, overlooking the fact that certain tokens contribute more to the overall reliability of a response.  
Recent work addresses this limitation by incorporating semantic awareness. Semantic entropy (SE)~\cite{kuhn2023semantic} reduces redundancy by grouping semantically equivalent outputs. Others re-weight token contributions\cite{duan2024shifting} or leverage hidden activations as uncertainty signals~\cite{liu2024uncertainty}. Building on this, we integrate dependency parse trees to identify key tokens shaping response meaning, while hidden activations provide semantic context. This combination enables a structured and context-aware approach to uncertainty estimation in LLMs.}

\noindent \textbf{Graph Pooling Approaches.} \textcolor{mycolor}{Graph pooling condenses input graphs while preserving key structural and semantic information. It generally falls into flat pooling, which applies simple aggregation functions like mean or sum~\cite{xu2019powerfulgraphneuralnetworks, 10.5555/2969442.2969488}, and hierarchical pooling, which progressively coarsens the graph to capture multi-level relationships~\cite{ying2018hierarchical}. Notable hierarchical methods include DiffPool~\cite{ying2018hierarchical}, which learns adaptive pooling assignments, and StructPool~\cite{yuan2020structpool}, which incorporates high-order structural dependencies. Other strategies include memory-based pooling~\cite{khasahmadi2020memorybasedgraphnetworks}, spectral filtering~\cite{defferrard2016convolutional}, and expressive pooling architectures~\cite{bianchi2023expressivepowerpoolinggraph}. Unsupervised pooling techniques, like mutual information maximization~\cite{10.1145/3511808.3557485}, further enable structure-preserving and label-free compression. This work proposes a hierarchical pooling approach leveraging dependency tree structures to improve uncertainty estimation. By representing LLM outputs as dependency graphs, \method captures both semantic and structural relationships, prioritizing key tokens for a more accurate and interpretable uncertainty assessment.}

\section{Background}
This section defines the problem, provides the necessary background, and features helpful for uncertainty estimation in LLMs, laying the foundation for our proposed approach.
\subsection{Problem Setup}
Uncertainty quantification in LLMs involves assessing confidence in LLM-generated responses based on input prompts. Given a prompt $\mathbf{x} = \{x_1, x_2, ..., x_k\}$, an LLM generates an output sequence $\mathbf{y} = \{y_1, y_2, ..., y_n\}$, where each token $y_j$ is sampled from a probability distribution conditioned on the prompt and prior tokens:
\begin{equation} 
    y_j \sim  p_{\theta}(\cdot|\mathbf{x},y_1,y_2,...,y_{j-1}),
\end{equation}
where $p_{\theta}$ represents the model's learned parameters. This next-token probability reflects how likely the model is to generate a particular token given the preceding context. Following~\cite{liu2024uncertainty}, \textcolor{mycolor}{when there is a downstream task, such as question answering or machine translation, a scoring function is introduced to evaluate the quality of the generated output. For such kinds of evaluation functions, factual truth or humans usually decide the true response. Thus the uncertainty estimation task can be} framed as a function $g(\mathbf{x}, \mathbf{y})$ that predicts the expected correctness of a response:
\begin{equation}
    g(\mathbf{x}, \mathbf{y}) \approx \mathbb{E}\left[s(\mathbf{y},\mathbf{y}_{\text{true}})|\mathbf{x}, \mathbf{y}\right].
\end{equation}
Here, $s(\mathbf{y}, \mathbf{y}_{\text{true}})$ denotes an evaluation metric comparing the generated response $\mathbf{y}$ with a ground-truth reference $\mathbf{y}_{\text{true}}$. The expectation is taken considering the semantic flexibility of natural language. The uncertainty arises from the input prompt $\mathbf{x}$ and the LLM itself rather than from a single absolute reference answer.

\subsection{Dependency Parse Trees in NLP}
Dependency parse trees provide a structured representation of syntactic relationships, defining hierarchical dependencies such as \emph{subjects}, \emph{objects}, and \emph{modifiers} within a sentence. These structures have been widely applied in various NLP tasks, including relation extraction (RE)~\cite{fundel2006relex, bjorne2009extracting}, named entity recognition (NER)~\cite{jie2017efficient}, and semantic role labeling (SRL)~\cite{marcheggiani-titov-2017-encoding}. They also enhance summarization by prioritizing salient information while filtering redundant content~\cite{li-etal-2014-improving, xu-durrett-2019-neural}. This work uses dependency parse trees to model structural relationships in LLM-generated text. These trees serve two key purposes: (1) They provide a hierarchical organization of tokens, helping distinguish pivotal words that shape response meaning, (2) They offer a consistent structure across different sentence formations, making them adaptable for modeling uncertainty in diverse LLM outputs.

\subsection{Features for Uncertainty Estimation}
\label{uncertainty_features}
Uncertainty estimation in LLMs relies on extracting meaningful features from the generated text. Prior studies~\cite{xiao-etal-2022-uncertainty, kadavath2022languagemodelsmostlyknow, lin2022teachingmodelsexpressuncertainty, tian-etal-2023-just, kuhn2023semantic, liu2024uncertainty} have demonstrated the effectiveness of token-level probability metrics. We categorize these features based on their sources~\cite{liu2024uncertainty}:

\noindent \textbf{White-box features:} These features are derived from hidden-layer activations, capturing the internal representation of tokens and providing insights into model confidence. These features are available only in open-source LLMs.

\noindent \textbf{Grey-box features:} These include \textit{token probabilities} and transformations such as entropy, offering uncertainty signals applicable to both open-source and commercial LLMs. The entropy of a discrete distribution $p$ over the vocabulary $\mathcal{V}$ is defined as $H(p) = -\sum_{v\in\mathcal{V}} p(v)\log\left(p(v)\right)$.
Given a prompt-response pair $(\mathbf{x},\mathbf{y})=(x_1,...,x_k,y_1,...,y_n)$, the entropy features for the $j$-th output token are given by $H(q_{\theta}(y_j|\mathbf{x},y_1,...,y_{j-1}))$, where $q_{\theta}$ denotes the LLM. The detailed mathematical definition of the features is provided in Appendix~\ref{ap: features description}.
\section{Approach}
This section details our approach, including graph formulation, hierarchical learning, and joint optimization, enabling a more structured and context-aware uncertainty estimation for LLMs.
\begin{figure}[t!]
    \centering
    \includegraphics[width=\linewidth]{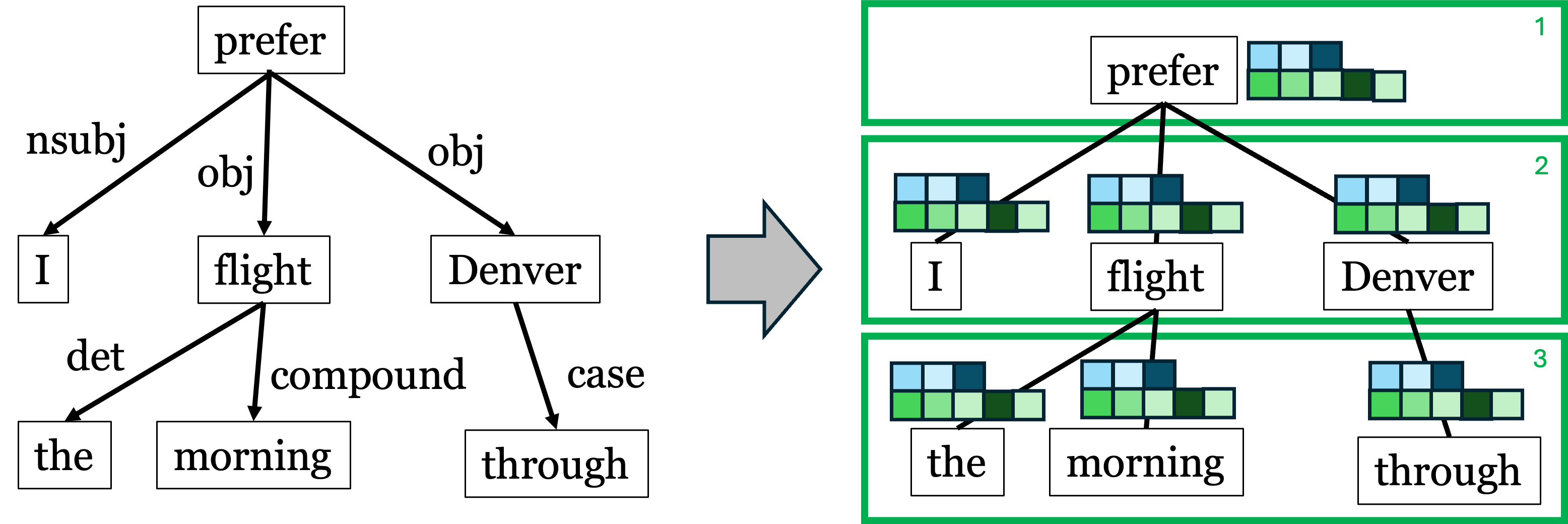}
    \caption{Dependency parse tree example. Each tree node is one token from the output. If two tokens have a relation, they are connected. Each tree node has additional features, such as probability, entropy, and embeddings (only for open-box LLMs).}
    \label{fig:parsing tree example}
    \vspace{-2pt}
\end{figure}
\subsection{Graph Formulation}
We transform dependency parse trees into graphs to structure LLM-generated text for uncertainty estimation. We first obtain the dependency tree using the Stanford NLTK parser, where each word serves as a node, and directed edges represent dependency relations. As shown in Fig.~\ref{fig:parsing tree example}, the root word, such as "prefer," has dependent words like "I" and "flight," forming a tree-like structure.

To extend this formulation beyond individual sentences, we construct a paragraph-level graph by linking the root nodes of multiple sentence-level dependency trees. Prior work~\cite{duan2024shifting} estimates uncertainty at the sentence level using a separate model to compute similarity, but such approaches may overlook deeper semantic relationships between sentences. Instead, \method learns inter-sentence relations directly, ensuring a more cohesive uncertainty estimation. Connecting root nodes across sentences enables cross-sentence token interactions, allowing uncertainty information to propagate effectively across the entire output. This formulation ensures that pivotal words influence the overall confidence estimation. The resulting global dependency graph provides a structured representation of LLM output, enhancing the ability of the proposed approach to assess uncertainty in LLM-generated text.

\subsection{Hierarchical Learning}
Transforming dependency parse trees into graphs enables us to frame uncertainty estimation as a graph aggregation problem, where each LLM-generated output is represented as a graph with nodes corresponding to words and edges capturing dependency relations. Each node has token-level features, such as next-token probability, entropy, and hidden state embeddings. We propose a hierarchical graph pooling approach inspired by semantic parsing trees~\cite{song2022hierarchical} to aggregate this information efficiently.

In a dependency graph (Fig.~\ref{fig:parsing tree example}), words appear at different levels based on their distance from the root token, which often signifies their semantic importance. Higher-level words generally play a more critical role in defining the sentence's meaning and, consequently, have a greater impact on uncertainty. To capture this, we introduce graph pooling, which groups tokens at different hierarchical levels, mitigating the effect of noisy words while assigning appropriate contributions to each token's uncertainty estimate.

Formally, given a dependency graph $\mathcal{G}=(\mathcal{V}, \mathcal{E})$, where $\mathcal{V}$ represents words and $\mathcal{E}$ defines their syntactic relations, we define an adjacency matrix $\mathcal{A} \in \mathbb{R}^{n \times n}$ and a feature matrix $\mathcal{X} \in \mathbb{R}^{n \times d}$. Inspired by hierarchical graph pooling methods~\cite{ying2018hierarchical}, we define the node clustering process using a learned soft assignment matrix:
\begin{equation}
    \mathcal{S}^l=\text{Softmax}(f(\mathcal{A}^l,\mathcal{X}^l, \theta_s)),
\end{equation}
where $\mathcal{A}^l$ and $\mathcal{X}^l$ represent the adjacency and feature matrices at pooling layer $l$, and $f$ in a GNN with learnable parameters $\theta_s$.

Before pooling, information propagates across the graph to model connectivity between clusters:
\begin{equation}
    \mathcal{Z}^l=f(\mathcal{A}^l,\mathcal{X}^l, \theta_z),
\end{equation}
where $\theta_z$ are the parameters of the GNN responsible for feature transformation. Using the learned assignment matrix $\mathcal{S}^l$, the graph is iteratively coarsened to generate a more compact representation:
\begin{equation} \label{eq: new graph from assignment}
    \begin{split}
        &\mathcal{X}^{l+1}=\mathcal{S}^l\mathcal{Z}^l \in \mathbb{R}^{n_{l+1}\times d}, \\
        &\mathcal{A}^{l+1}=\mathcal{S}^l\mathcal{A}^l\mathcal{S}^{l^T} \in \mathbb{R}^{n_{l+1}\times n_{l+1}}.
    \end{split}
\end{equation}
Here, $\mathcal{X}^l$ and $\mathcal{A}^l$ are iteratively refined representations at each pooling level, ensuring that semantically important tokens retain greater influence. By hierarchically aggregating token-level uncertainty, \method enhances interpretability and robustness, providing a structured estimation of confidence in LLM-generated responses.
\begin{figure*}[ht]
    \centering
    \includegraphics[width=\textwidth]{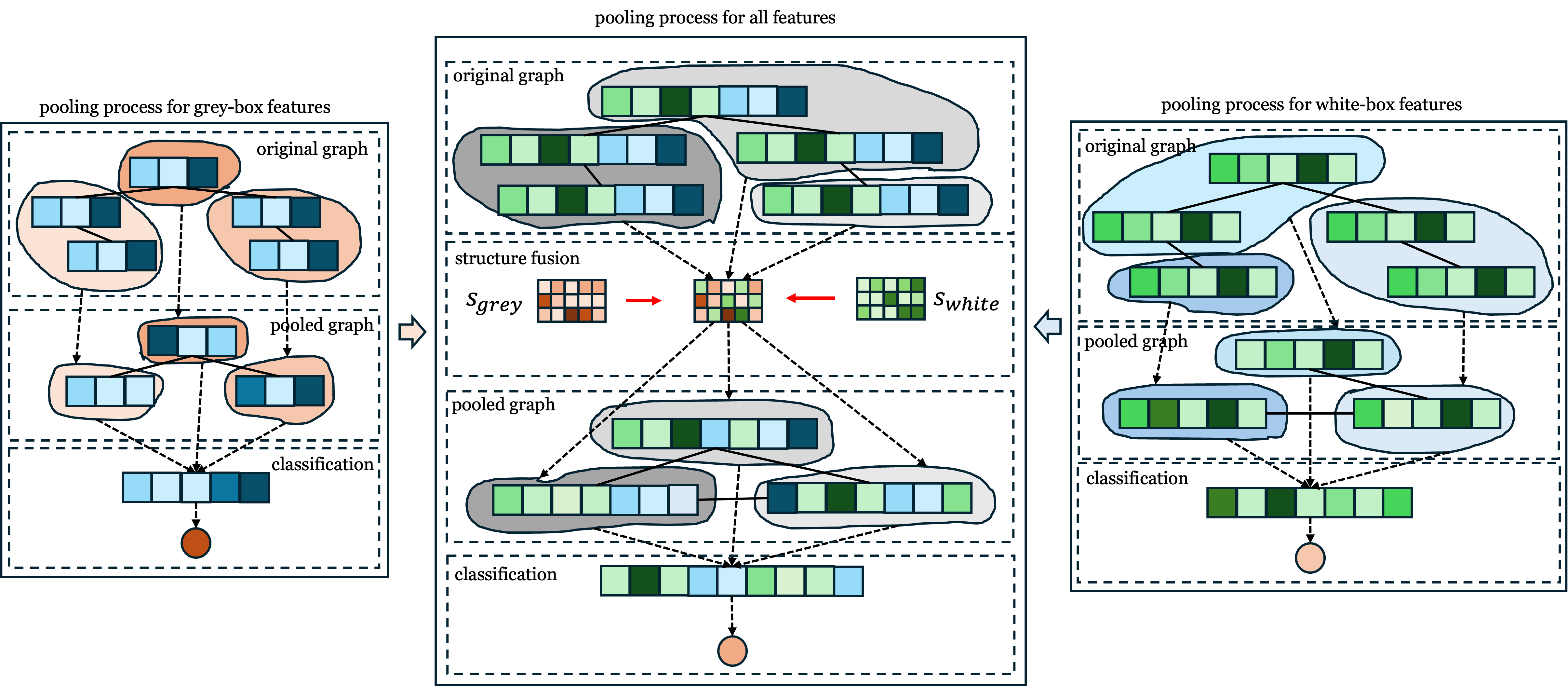}
    \caption{The Overview of \method, composed of three modules: (1) pooling based on grey-box features, (2) pooling based on white-box features, and (3) a learnable fusion process integrating both modules. Both the grey-box pooling process and white-box pooling process share the same graph structure, but differ in features, which leads to different fusion matrices. The structure fusion process helps better integrate various fusion matrices.}
    \label{fig:framework}
    \vspace{-1.5em}
\end{figure*}
\subsection{Joint Optimization}
Uncertainty estimation in LLMs relies on multiple features, as discussed in Section~\ref{uncertainty_features}, including hidden states (white-box features) and probability-based signals (grey-box features), each contributing differently. Prior work~\cite{liu2024uncertainty} shows that hidden states encode valuable uncertainty information, partly due to the misalignment between pretraining objectives and uncertainty estimation. Moreover, hidden states capture semantic relationships among tokens, making them especially important for confidence evaluation.

We propose a joint optimization framework to effectively integrate multiple uncertainty features. As illustrated in Fig.~\ref{fig:framework}, \method includes a semantic pooling module that leverages hidden state embeddings and a structural pooling module that utilizes probability and entropy features. Both modules operate on a shared dependency parse tree, providing a unified structural backbone. Their outputs are combined via a fusion module that learns a joint graph pooling matrix, balancing semantic and structural signals to refine uncertainty estimation. Instead of merging features at the node level, we fuse them at the assignment matrix level to better balance structural and semantic information. This design is motivated by three factors. First, direct feature fusion would bias toward embeddings due to their higher dimensionality. Second, embeddings encode semantic context but lack precise generation uncertainty, while probability and entropy features provide more accurate confidence signals. Third, the assignment matrix inherently reflects token importance and relational structure, making it a more effective fusion point for heterogeneous features.

To achieve this, we introduce an end-to-end learnable fusion module, where the fused assignment matrix is computed as:
\begin{equation}
    \mathcal{S}_*^l=\text{Softmax}(g(\mathcal{S}_{\text{grey}}^l, \mathcal{S}_{\text{white}}^l,\theta_{s*})),
\end{equation}
where $\mathcal{S}_{\text{grey}}^l$ and $\mathcal{S}_{\text{white}}^l$ are the assignment matrices at pooling layer $l$ from the structural and semantic modules, respectively, and $\theta_{s*}$ denotes the learnable parameters of the fusion function $g$.

Following this, a GNN propagates information across the graph, refining node representations:
\begin{equation}
    \mathcal{Z}_*^l=f(\mathcal{A}_*^l, \mathcal{X}_*^l,\theta_{z*}),
\end{equation}
where $f$ is a GNN with learnable parameters $\theta_{z*}$. These updated assignment and node embedding matrices are used to refine the graph iteratively:
\begin{equation}
    \begin{split}
        &\mathcal{X}_*^{l+1}=\mathcal{S}_*^l\mathcal{Z}_*^l, \\
        &\mathcal{A}_*^{l+1}=\mathcal{S}_*^l\mathcal{A}_*^l\mathcal{S}_*^{l^T}.
    \end{split}
\end{equation}
Here, $\mathcal{X}_*$ encodes probability and entropy features, while embeddings enhance the model's semantic understanding. The independent assignment matrices $\mathcal{S}_{\text{grey}}^l$ and $\mathcal{S}_{\text{white}}^l$ are jointly optimized to capture both structural and contextual uncertainty, improving the robustness of LLM confidence evaluation. \textcolor{mycolor}{Recall that when using open-box LLMs, which allow users access to grey-box features and white-box features, our fusion process can be directly applied. While using black-box LLMs, which only allow access to grey-box features, the fusion process can not proceed without white-box features. However, this will not hinder the application of \method as the graph structures and the joint optimization will remain effective in estimating the uncertainty. We provide both the results with only grey-box features and both grey-box and white-box features in the experiments in Fig.~\ref{fig: experimental results} to prove.}

\section{Experiments}
This section evaluates \method across multiple dimensions: (1) effectiveness in assessing uncertainty (Section~\ref{sec: quantitative analysis}), (2) an ablation study to analyze the role of two modules (Section~\ref{sec: alabtion study}), (3) a scalability test to assess computational efficiency (Section~\ref{sec: scalability test}), (4) the impact of dependency parse trees on uncertainty estimation (Appendix~\ref{sec: graph structure comparison}), (5) a parameter analysis to determine the sensitivity of \method to hyperparameter tuning (Appendix~\ref{sec: parameter analysis}), (6) the impact of LLM parameters on \method's uncertainty estimation performance (Appendix~\ref{sec: parameter impact}), and (7) the impact of training dataset size and noisy labels on \method's performance (Section~\ref{sec: robustness test} and Appendix \ref{ap: section robustness test}). Due to space constraints, the results on dimensions 4, 5, 6, and 7 are presented in Appendix~\ref{ap: additional experiments}.

\subsection{Experimental Setup}
We evaluate \method using different LLM architectures, multiple datasets spanning various NLP tasks, and state-of-the-art baselines.
\begin{figure*}[ht!]
    \centering
    \includegraphics[width=\linewidth]{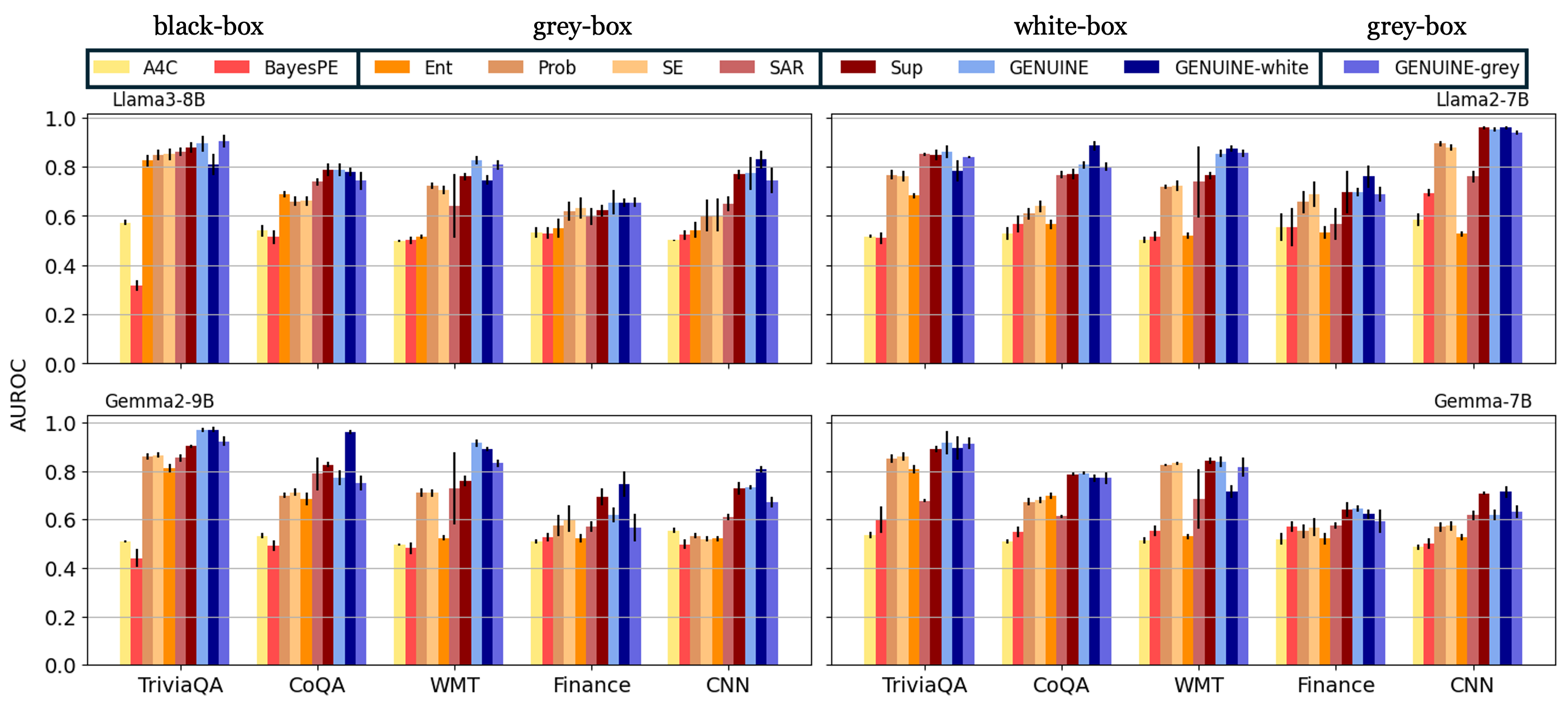}
    \caption{Comparison of AUROC on five datasets, four LLMs,  and seven baselines. Error bars denote variance over five runs. \method and its transformations outperform other baselines for all five datasets and four LLMs. Especially for datasets with relatively longer output from LLMs, such as WMT, Finance, and CNN datasets}
    \label{fig: experimental results}
    \vspace{-1.5em}
\end{figure*}

\noindent \textbf{LLMs.} We consider open-source LLMs, including Llama2-7B, Llama2-13B, Llama3-8B~\cite{touvron2023llama}, as well as Gemma-7B and Gemma2-9B~\cite{gemma_2024}. The respective tokenizers provided by Hugging Face are used, and model parameters remain unchanged.

\noindent \textbf{Datasets.} We evaluate the uncertainty estimation on three NLP tasks: question answering (CoQA\cite{reddy-etal-2019-coqa}, TriviaQA \cite{joshi-etal-2017-triviaqa}, and Finance QA dataset~\cite{alpaca}), machine translation (WMT 2014 dataset \citep{bojar2014findings}), and summarization (CNN dataset~\cite{hermann2015teaching}). The details of the datasets are introduced in Appendix~\ref{ap: dataset introduction}. Each dataset is split into training (60\%), validation (10\%), and test (30\%) sets, with five runs performed to mitigate the effects of randomness in parameter optimization. Few-shot prompting is adopted, with templates detailed in Appendix~\ref{ap: prompt template}.

\noindent \textbf{Baselines.} We include five categories of state-of-the-art baselines to compare \method against with: (1) A4C~\cite{tian2023just}, which directly queries the LLM for its self-assessed uncertainty, (2) Entropy and probability-based methods, including Avg Probability (Prob) and Avg Entropy (Ent), as defined in Table~\ref{tab: features definition} in the Appendix \ref{ap: features description}, (3) Semantic-aware methods, such as Semantic Entropy (SE)~\cite{kuhn2023semantic} and SAR~\cite{duan2024shifting}, (4) Bayesian based methods, including BayesPE\cite{tonolini-etal-2024-bayesian}, and (5) A supervised uncertainty estimation approach (Sup)~\cite{liu2024uncertainty}.
Details of the prompt templates are provided in the Appendix~\ref{ap: prompt template}.

\noindent \textbf{Evaluation Metrics.} Following~\cite{liu2024uncertainty, kuhn2023semantic}, we evaluate \method’s ability to distinguish correct from incorrect responses using uncertainty scores. Our primary metric is AUROC, which measures how well the model ranks correct responses above incorrect ones. We also assess calibration using Expected Calibration Error (ECE)\cite{naeini2015obtaining}, and report Brier score\cite{hernandez2011brier} and negative log-likelihood (NLL)\cite{hastie01statisticallearning} to evaluate the alignment between predicted uncertainty and true confidence. AUROC is shown in the main paper, others are reported in Appendix~\ref{calibration_performance}.

\subsection{Performance of Uncertainty Estimation} \label{sec: quantitative analysis}
We evaluate \method using the AUROC metric against state-of-the-art baselines. As shown in Fig.~\ref{fig: experimental results}, \method consistently outperforms prior methods, particularly on long-form generation tasks (WMT, Finance, CNN). Its dependency-based structural modeling improves uncertainty estimation by reducing error propagation across extended sequences. \method also achieves better calibration, as evidenced by lower ECE, NLL, and Brier scores (Appendix~\ref{calibration_performance}), minimizing uncertainty misalignment in downstream tasks. The results further highlight that response length significantly impacts uncertainty estimation. As detailed in Table~\ref{tab: graphs statistics}, \method offers modest AUROC gains on shorter outputs (e.g., TriviaQA, CoQA), but shows substantial improvements on longer responses (e.g., WMT, Finance, CNN). Traditional token-wise methods accumulate errors over extended text, whereas \method’s structured approach better handles long-form content, critical for tasks like dialogue and summarization.

\noindent Feature selection also plays a crucial role in uncertainty estimation. While combining multiple features generally improves performance, hidden-layer embeddings alone (\method-white) perform best on Finance and CNN datasets, where longer sequences amplify token-level error in entropy-based methods. 
\textcolor{mycolor}{To support both black-box and open-box LLMs, we introduce two variants: \method-grey (using only grey-box features) and \method-white (white-box features). The results demonstrate that in most cases, \method-grey still has the superiority of performance, which shows the applicability of \method in black-box LLMs.}
These findings also highlight the advantage of open-source LLMs with access to internal representations for robust uncertainty modeling.


\subsection{Ablation Study} \label{sec: alabtion study}

\textcolor{mycolor}{\method introduces a graph structure and fusion mechanism to balance structural and semantic information. We conduct ablation studies on the TriviaQA dataset using Llama3-8B, Llama2-7B, Gemma2-9B, and Gemma-7B to assess the contribution of each component. Due to space constraints, we report results for Llama3-8B and Gemma2-9B, with full results in Appendix~\ref{ap: alabtion study}. We denote variants as \method w/o fusion \& graph (without both modules) and \method w/o fusion (with graph, but no fusion). As shown in Table~\ref{tab: ablation on fusion process}, the graph structure and the fusion process improve AUROC on all the LLMs we use in our experiments. These findings highlight the graph structure and the fusion strategy’s effectiveness in integrating structural and semantic signals, enabling better uncertainty propagation. We observe that the improvement in Gemma2-9B model is not significant. The smaller gains for Gemma2-9B may be due to its already strong baseline performance (AUROC > 0.95), leaving limited room for improvement.}
\begin{table}[ht!]
    \centering
    \small
    \caption{Ablation study on TriviaQA. \method w/o fusion \& graph (without both modules) and \method w/o fusion (with graph, but no fusion) ($\uparrow$ means the higher the better)}
    \begin{tabular}{c|c|c}
    \hline
        \multicolumn{3}{c}{TriviaQA} \\ \hline
        \multirow{2}{*}{Methods} & Llama3-8B & Gemma2-9B \\
        \cline{2-3}
        ~ & AUROC $\uparrow$ & AUROC $\uparrow$ \\ \hline
        \makecell{\method w/o \\ fusion \& graph} & 0.789±0.031 & 0.956±0.002 \\
        \method w/o fusion & 0.809±0.096 & 0.963±0.015 \\ 
        \method & \textcolor{darkpastelgreen}{0.894±0.032} & \textcolor{darkpastelgreen}{0.969±0.009} \\
        \hline
        \hline
        \multicolumn{3}{c}{WMT} \\ \hline
        \multirow{2}{*}{Methods} & Llama3-8B & Gemma2-9B \\
        \cline{2-3}
        ~ & AUROC $\uparrow$ & AUROC $\uparrow$ \\ \hline
        \makecell{\method w/o \\ fusion \& graph} & 0.709±0.013 & 0.898±0.012 \\
        \method w/o fusion & 0.713±0.002 & 0.905±0.002 \\ 
        \method & \textcolor{darkpastelgreen}{0.826±0.019} & \textcolor{darkpastelgreen}{0.914±0.014} \\
        \hline
    \end{tabular}
    \label{tab: ablation on fusion process}
\end{table}

\subsection{Scalability} \label{sec: scalability test}
\begin{figure}[t!]
    \centering
    \begin{subfigure}[t]{0.47\linewidth}
        \includegraphics[width=\textwidth]{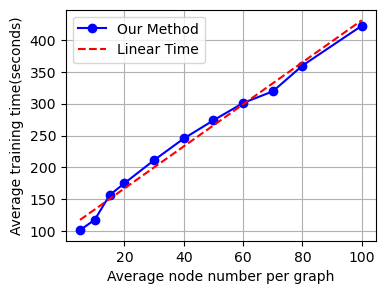}
        \caption{Scalability test on the number of nodes per graph.}
        \label{fig:scalability number of nodes}
    \end{subfigure}
    ~
    \begin{subfigure}[t]{0.48\linewidth}
        \includegraphics[width=\textwidth]{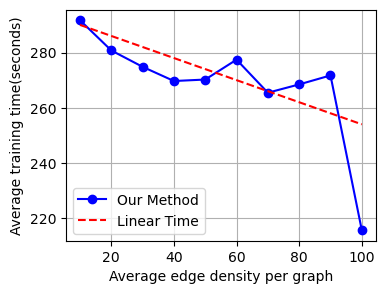}
        \caption{Scalability test on graph density.}
        \label{fig:scalability edge density}
    \end{subfigure}
    \caption{Scalability test on the node number and edge density}
\end{figure}

\textcolor{mycolor}{We evaluate the scalability of \method by examining its computational efficiency with increasing node count and graph density. As shown in Fig.~\ref{fig:scalability number of nodes}, training time scales near-linearly with the number of nodes, demonstrating that \method remains computationally feasible even for larger graphs. This suggests that the model can efficiently process uncertainty in large-scale LLM outputs without excessive overhead. In Fig.~\ref{fig:scalability edge density}, computational cost decreases as graph density increases, indicating that denser graphs facilitate more efficient uncertainty aggregation. Sparse graphs (e.g., 10\% density) require 1.5 times more processing time than fully connected graphs (100\% density), emphasizing the trade-off between structure complexity and efficiency. These findings confirm that \method scales effectively with increasing graph complexity, making it well-suited for high-dimensional NLP tasks such as document summarization, multi-turn dialogue, and knowledge-intensive reasoning. Its ability to maintain efficiency while capturing semantic and structural relationships ensures its adaptability to real-world LLM evaluation scenarios.}

\subsection{Robustness} \label{sec: robustness test}
\textcolor{mycolor}{In real-world scenarios, uncertainty estimation models often face limited training data and noisy labels, which can affect performance. To evaluate the robustness of \method under such conditions, we conduct experiments using the Llama3-8B model on the TriviaQA dataset. Table~\ref{ap: training size impact on performance} in the Appendix~\ref{ap: section robustness test} shows how varying training set sizes impact performance. Please refer to the Appendix for more details. While Table~\ref{ap: noisy label impact on performance} examines the effect of label noise. For the latter, we randomly corrupt a portion of training labels (as specified by the noise ratio) and assess performance on the clean test set. Specifically, Table~\ref{ap: noisy label impact on performance} shows that label noise negatively affects model performance. But \method remains robust when up to 0.1\% of the training labels are corrupted. However, AUROC declines sharply when the noise ratio increases, and so do the calibration metrics. These experiments demonstrate \method’s resilience to data scarcity and label noise, highlighting its applicability in real-world settings.}
\begin{table}[ht!]
    \centering
    \small
    \caption{\textcolor{mycolor}{The impact of noisy labels on \method performance. \method remains robust with 0.1\% of labels being noisy.($\uparrow$ means the higher the better, $\downarrow$ means the lower the better)}}
    \begin{tabular}{c|c|c}
    \hline
        noise ratio & AUROC$\uparrow$ & ECE$\downarrow$\\ \hline
        0 & 0.894±0.032 & 0.246±0.007 \\ 
        0.001 & 0.894±0.017 & 0.244±0.010 \\ 
        0.003 & 0.863±0.033 & 0.243±0.009  \\ 
        0.005 & 0.855±0.024 & 0.243±0.013 \\ 
        0.01 & 0.821±0.014 & 0.240±0.014 \\ 
        0.02 & 0.705±0.037 & 0.235±0.015 \\ 
        0.03 & 0.746±0.095 & 0.232±0.019 \\ 
        0.04 & 0.672±0.140 & 0.234±0.022 \\ \hline
        \hline
        noise ratio & NLL$\downarrow$ & Brier$\downarrow$ \\ \hline
        0 & 0.362±0.005 & 0.094±0.002 \\ 
        0.001 & 0.364±0.005 & 0.095±0.002 \\ 
        0.003 & 0.366±0.002 & 0.096±0.001 \\ 
        0.005 & 0.370±0.008 & 0.098±0.004 \\ 
        0.01 & 0.377±0.009 & 0.101±0.004 \\ 
        0.02 & 0.390±0.019 & 0.107±0.009 \\ 
        0.03 & 0.407±0.026 & 0.114±0.012 \\ 
        0.04 & 0.414±0.031 & 0.117±0.014 \\ \hline
    \end{tabular}
    \label{ap: noisy label impact on performance}
\end{table}

\subsection{Graph Structure and Uncertainty Estimation} \label{sec: graph structure comparison}
\noindent Understanding the impact of graph structure on uncertainty estimation is essential for refining confidence assessment in LLM-generated responses. This section evaluates the effectiveness of dependency parse trees and analyzes graph structure variations across datasets and LLMs, using results from Table~\ref{tab:graph comparison} and Table~\ref{tab: graphs statistics}.

\noindent \textbf{Dependency Parse Trees vs. Next-Token Graphs.} To assess the impact of different graph structures, we compare the dependency parse tree (DPT) against the next-token graph (NTG), where edges only connect adjacent words in a sentence. The results in Table~\ref{tab:graph comparison} clearly demonstrate that DPT-based graphs consistently outperform NTG-based graphs across all evaluation metrics, reinforcing the importance of semantic structure in uncertainty estimation.

For Llama3-8B, DPT achieves an AUROC of 0.894, improving over NTG (0.885), while also achieving lower ECE (0.246 vs. 0.264), NLL (0.362 vs. 0.437), and Brier score (0.094 vs. 0.130). Similar trends hold for Gemma2-9B, where DPT significantly outperforms NTG with an AUROC improvement of nearly 6\% (0.905 vs. 0.846) and lower calibration errors. These results confirm that structural relationships encoded in dependency graphs improve uncertainty estimation, providing richer contextual information than simple word adjacency models.

When comparing grey-box and white-box features, we observe that DPT consistently performs better than NTG in both settings. For instance, DPT w/ grey achieves an AUROC of 0.903 for Llama3-8B, outperforming NTG w/ grey (0.897) while maintaining better calibration across ECE, NLL, and Brier scores. The trend holds for white-box features, where DPT w/ white achieves better AUROC than NTG w/ white, showing that dependency parsing enhances uncertainty modeling even when using only hidden-layer embeddings.

Besides model performance comparison, we also provide some qualitative examples to further demonstrate the effectiveness of our method. In Table \ref{tab: case study}, we provide three examples randomly selected from the TriviaQA dataset. In the table, we offer the ground truth, the answers generated from LLMs, the adjacency matrix, and the output probability. These examples show that NTG's reliance on sequential adjacency fails to generalize, especially when the position of key tokens changes or the sentence structure becomes more complex. In contrast, DPT maintains high accuracy by leveraging semantic roles derived from syntax, regardless of token order.

\begin{table}[ht!]
    \centering
    \small
    \caption{Examples from the TriviaQA dataset between NTG and DPT.}
    \begin{tabular}{c|c}
    \hline
        question & \makecell{The Solovetsky Islands lie in \\ which body of water?} \\ \hline
        ground truth & The White Sea  \\ 
        LLM answer & The White Sea  \\ 
        adjacency matrix & [[0, 0, 1],[0, 0, 1], [1, 1, 0]]  \\ 
        final output (DPT) & [0.0013, 0.9987], correct  \\ 
        final output (NTG) & [0.1587, 0.8413], correct \\
        \hline
        \hline
        question & \makecell{Which writer had a pet cat \\ called Caterina, that was \\the inspiration for his story \\ ‘The Black Cat’?} \\ \hline
        ground truth & Edgar Allen Poe  \\ 
        LLM answer & Edgar Allen Poe  \\ 
        adjacency matrix & [[0, 1, 1],[1, 0, 0], [1, 0, 0]]  \\ 
        final output (DPT) & [0.3815, 0.6185], correct  \\ 
        final output (NTG) & [0.5407, 0.4593], wrong \\
        \hline
        \hline
        question & \makecell{What is the name of the hotel \\ in the 1980 film ‘The Shining’, \\ starring Jack Nicholson?} \\ \hline
        ground truth & The Overlook Hotel  \\ 
        LLM answer & The Overlook Hotel  \\ 
        adjacency matrix & [[0, 0, 1],[0, 0, 1], [1, 1, 0]]  \\ 
        final output (DPT) & [0.3765, 0.6235], correct  \\ 
        final output (NTG) & [0.5859, 0.4141], wrong \\
        \hline
    \end{tabular}
    \label{tab: case study}
\end{table}

These findings suggest that semantic-aware uncertainty estimation is essential, especially for longer text sequences where sequential token dependencies alone fail to capture structural nuances. By modeling hierarchical relations, DPT-based uncertainty estimation improves both reliability and calibration, making it particularly useful for structured prediction tasks.
\section{Conclusion}

This paper introduces dependency-based semantic structures for uncertainty estimation in LLMs. Our findings prove that incorporating structural information enhances uncertainty modeling, leading to more accurate and calibrated estimates. \method outperforms existing uncertainty estimation methods (AUROC), particularly in long-form text generation, while also improving calibration metrics (ECE, NLL, Brier). Our results show that semantic graphs derived from dependency parse trees enhance uncertainty modeling, making them valuable for evaluating LLMs' outputs and guiding future improvements in adaptive uncertainty estimation in dynamic, real-world settings.
\section{Ethical Consideration}
\method enhances the credibility and reliability of LLMs by improving uncertainty estimation, helping to mitigate the risks of misinformation. By refining confidence assessment, \method reduces misinformation and promotes more trustworthy AI-generated content.

However, several ethical limitations must be considered. Uncertainty estimation does not prevent misinformation but provides a measure of confidence, which still requires human interpretation. Over-reliance on uncertainty scores could lead to misjudgments, either overestimating or underestimating the reliability of LLM outputs. Additionally, \method's effectiveness depends on dependency parsing and feature selection, which may introduce biases if trained on imbalanced datasets. Furthermore, while \method improves model calibration, uncertainty quantification remains imperfect, and its reliability may vary across domains, particularly in high-stakes applications such as healthcare, finance, and law. Addressing these challenges requires ongoing evaluation, transparency, and responsible deployment to ensure ethical and fair AI use.
\section{Limitations}
\method introduces a graph-based approach for confidence evaluation in LLMs, but certain limitations remain. \method relies on token logits and embeddings, which, though widely available in open-source and commercial LLMs, may limit its applicability in black-box scenarios where such information is restricted. Additionally, its performance is influenced by generation length and labeled data availability, making it sensitive to dataset variability. Finally, this study focuses on NLP tasks and datasets, leaving open the exploration of its effectiveness in multimodal and cross-domain applications.
\bibliography{custom}

\appendix
\section*{Appendix}

\section{Implementation Details}
This section provides an overview of the implementation details of \method.
\subsection{Details of Datasets} \label{ap: dataset introduction}
Here in this section, we provide more details on the datasets.

\noindent \textit{Question Answering.} We use the CoQA~\cite{reddy-etal-2019-coqa} and TriviaQA~\cite{joshi-etal-2017-triviaqa} datasets to assess LLMs' ability to generate responses based on contextual understanding and pre-trained knowledge. Additionally, we include the Finance QA dataset~\cite{alpaca}, which evaluates domain-specific knowledge in financial contexts. Rouge-1~\cite{lin-och-2004-automatic} is used as the scoring function, labeling a response $\mathbf{y}_i$ as correct if $s(\mathbf{y}_i, \mathbf{y}_{i,\text{true}}) \geq 0.3$.

\noindent \textit{Machine Translation.} We evaluate translation quality using the WMT 2014 dataset~\cite{bojar2014findings}, with BLEU score~\cite{papineni2002bleu} as the metric. A response $\mathbf{y}_i$ is considered correct if $s(\mathbf{y}_i, \mathbf{y}_{i,\text{true}}) \geq 0.3$.


\noindent \textit{Summarization.} The CNN~\cite{hermann2015teaching} dataset is used for summarization task, where generated outputs are labeled as correct if they achieve a Rouge-L score of at least 0.35, following~\cite{quach2024conformallanguagemodeling}.

\subsection{Details of Features} \label{ap: features description}
This section provides the mathematical definitions of the features used in our uncertainty estimation framework. A detailed breakdown is presented in Table~\ref{tab: features definition}.
\begin{table}[h]
    \setlength{\tabcolsep}{0pt}
    \caption{Features used for the supervised task of uncertainty estimation for LLMs.}
    \begin{tabular}{c|cc}
        \hline
        Name & Definition \\
        \hline
        Ent & $H(p_\theta(\cdot|\mathbf{x},y_1,\ldots,y_{j-1}))$\\
        Max Ent & $\max_{j\in\{1,...,n\}}\ H(p_\theta(\cdot|\mathbf{x},y_1,\ldots,y_{j-1}))$ \\
        Min Ent & $\min_{j\in\{1,...,n\}}\ H(p_\theta(\cdot|\mathbf{x},y_1,\ldots,y_{j-1}))$ \\
        Avg Ent & $\frac{1}{n} \sum_{j=1}^n H(p_\theta(\cdot|\mathbf{x},y_1,\ldots,y_{j-1}))$ \\
        Std Ent & $\sqrt{\frac{\sum_{j=1}^n\left(H(p_\theta(\cdot|\mathbf{x},y_1,\ldots,y_{j-1}))-\text{Avg Ent}\right)^2}{n-1}}$ \\
        Prob & $p_{\theta}(y_j|\mathbf{x},y_1,\ldots,y_{j-1})$ \\
        Max Prob & $\max_{j\in\{1,...,n\}}\ p_{\theta}(y_j|\mathbf{x},y_1,\ldots,y_{j-1})$ \\
        Min Prob & $\min_{j\in\{1,...,n\}}\ p_{\theta}(y_j|\mathbf{x},y_1,\ldots,y_{j-1})$ \\
        Avg Prob & $\frac{1}{n}\sum_{j=1}^n p_{\theta}(y_j|\mathbf{x},y_1,\ldots,y_{j-1})$ \\
        Std Prob & $\sqrt{\frac{\sum_{j=1}^n\left(p_{\theta}(y_j|\mathbf{x},y_1,\ldots,y_{j-1})-\text{Avg Prob}\right)^2}{n-1}}$ \\
        \hline
    \end{tabular}
    \label{tab: features definition}
\end{table}

\subsection{Prompt Template} \label{ap: prompt template}
We adopt a few-shot prompting strategy, following the approach of \cite{liu2024uncertainty}. Each prompt comprises four components: introduction, examples, question, and answer. The examples are user-defined question-answer pairs structured identically to the target task, ensuring consistency in format. The model receives the formatted template along with the reference question and is prompted to generate an appropriate response. This structured approach helps standardize uncertainty estimation across different tasks.
\begin{center}
    \fcolorbox{black}{gray!10}{
        \parbox{\linewidth}{
            \textbf{TriviaQA}
            
            \texttt{Answer the question as following examples.}
            \texttt{Examples: \textit{Q: What star sign is Michael Caine? A: Pisces. Q: Which George invented the Kodak roll-film camera? A: Eastman. Q: ... A: ...}}
            
            \texttt{Q: In which decade was Arnold Schwarzenegger born?}
            \texttt{A: 1950s}
        }
    }
\end{center}

\begin{center}
\fcolorbox{black}{gray!10}{
    \parbox{\linewidth}{
        \textbf{CoQA}
        
        \texttt{Reading the passage and answer given questions accordingly.}
        \texttt{Passage: The Vatican Apostolic Library, more commonly called the Vatican Library or simply the Vat, is the library of the Holy See, located in Vatican City. ...}
        \texttt{Examples: }
        \texttt{Q: When was the Vat formally opened? A: It was formally established in 1475. Q: ... A: ...}
        
        \texttt{Q: what was started in 2014?}
        \texttt{A: a project.}
        }
    }
\end{center}

\begin{center}
    \fcolorbox{black}{gray!10}{
        \parbox{\linewidth}{
            \textbf{WMT}
            
            \texttt{What is the English translation of the following sentence? }
            \texttt{Q: Spectaculaire saut en \"wingsuit\" au-dessus de Bogota. A: Spectacular Wingsuit Jump Over Bogota. Q: ... A: ...}
            
            \texttt{Q: Une boîte noire dans votre voiture ?}
            \texttt{A: A black box in your car?}
        }
    }
\end{center}

\begin{center}
    \fcolorbox{black}{gray!10}{
        \parbox{\linewidth}{
            \textbf{Finance}
            
            \texttt{Answer the question as following examples.}
            \texttt{Examples: \textit{Q: For a car, what scams can be plotted with 0\% financing vs rebate? A: he car deal makes money 3 ways. If you pay in one lump payment. ... Q: ... A: ...}}
            
            \texttt{Q: Where should I be investing my money?}
            \texttt{A: Pay off your debt. As you witnessed, no "investment" \% is guaranteed. ...}
        }
    }
\end{center}

\begin{center}
    \fcolorbox{black}{gray!10}{
        \parbox{\linewidth}{
            \textbf{Finance}
            
            \texttt{What are the highlights in this paragraph?}
            \texttt{Examples: \textit{Q: LONDON, England (Reuters) -- Harry Potter star Daniel Radcliffe gains access to a reported £20 million (\$41.1 million) fortune ... A: Harry Potter star Daniel Radcliffe gets £20M fortune as he turns 18 Monday . ... Q: ... A: ...}}
            
            \texttt{Q: Editor's note: In our Behind the Scenes series, CNN correspondents share ...}
            \texttt{A: Mentally ill inmates in Miami are housed on the "forgotten floor" ...}
        }
    }
\end{center}

\section{Additional Experiments}
\label{ap: additional experiments}
\begin{figure*}
    \centering
    \includegraphics[width=\linewidth]{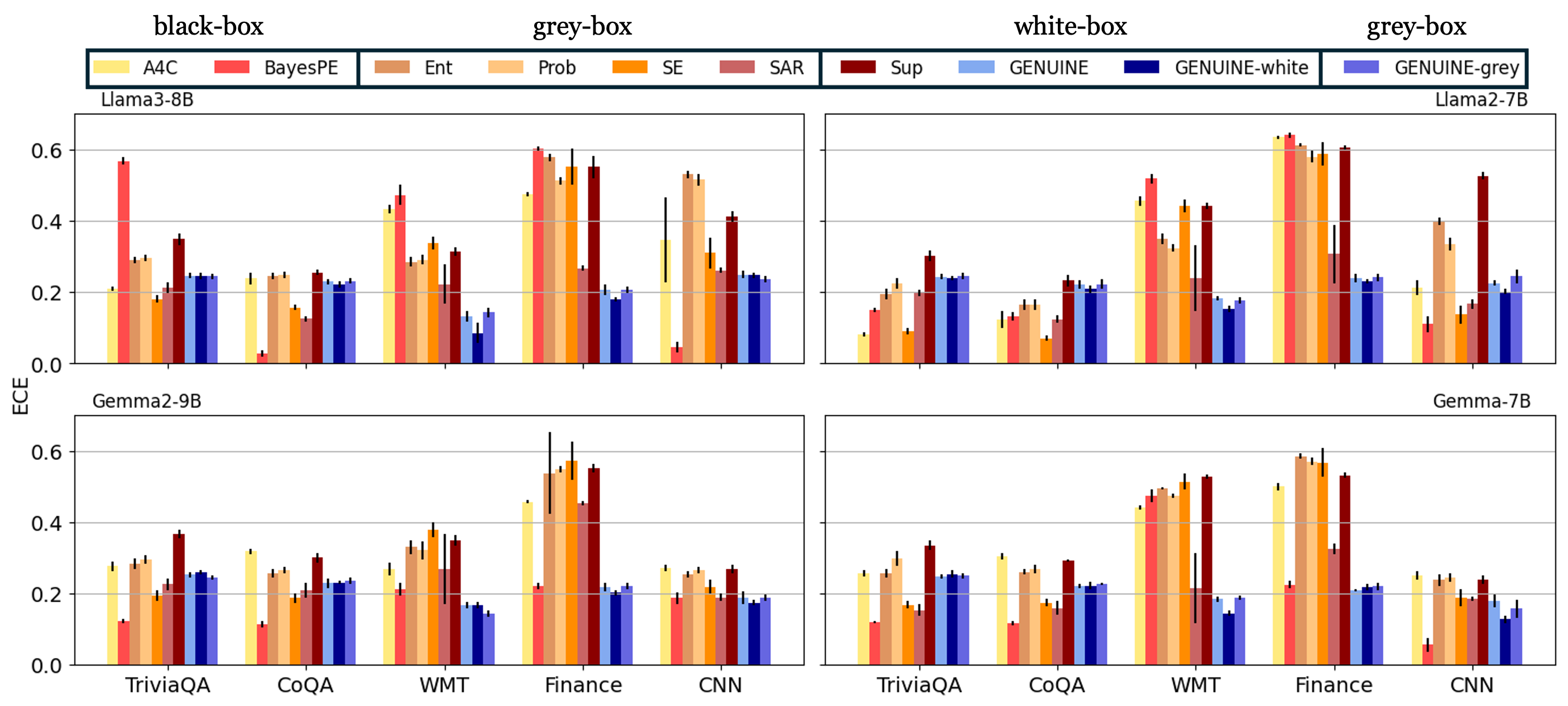}
    \caption{Comparison of ECE on five datasets, four LLMs, and seven baselines. Error bars denote variance over five runs.}
    \label{fig: ece results}
\end{figure*}

\begin{figure*}
    \centering
    \includegraphics[width=\linewidth]{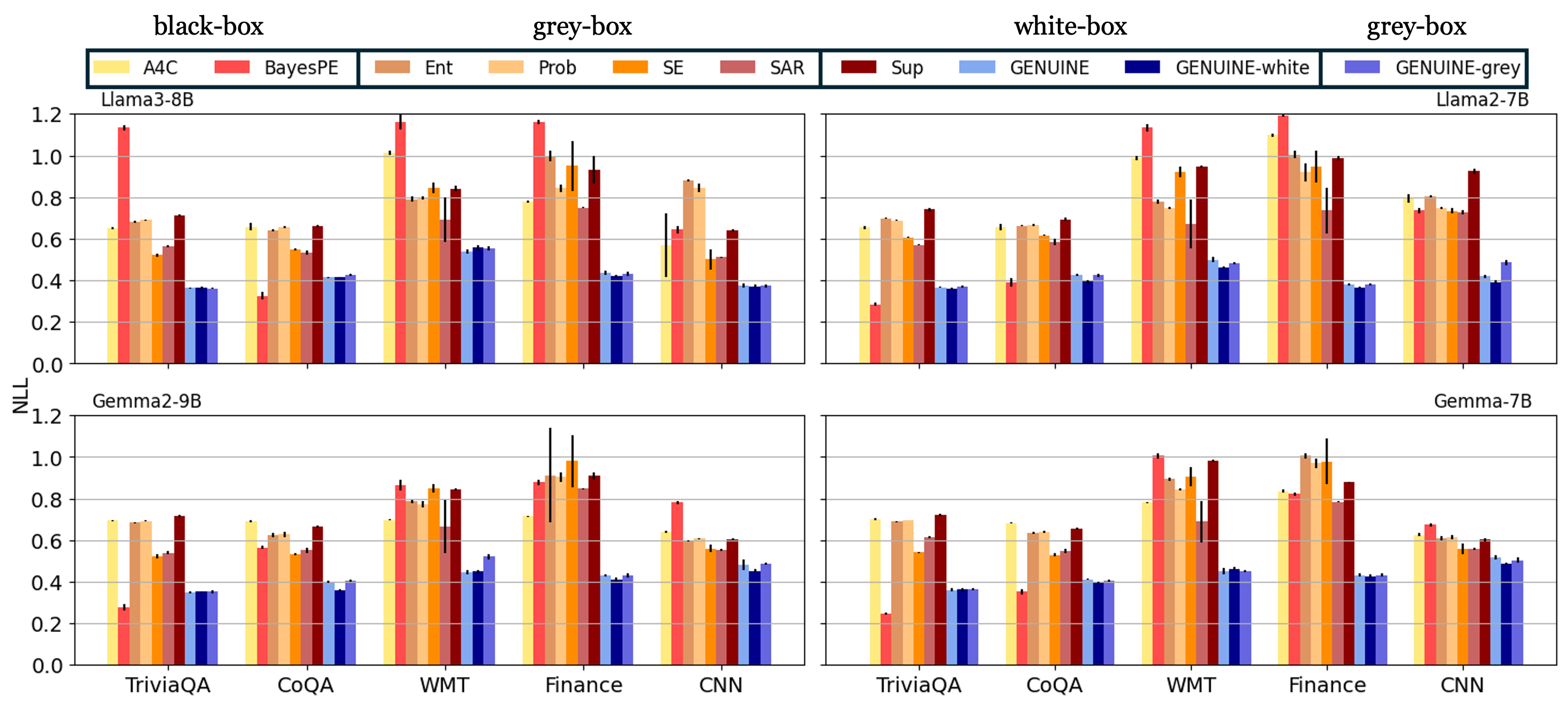}
    \caption{Comparison of NLL on five datasets, four LLMs, and seven baselines. Error bars denote variance over five runs.}
    \label{fig: nll results}
\end{figure*}

\begin{figure*}
    \centering
    \includegraphics[width=\linewidth]{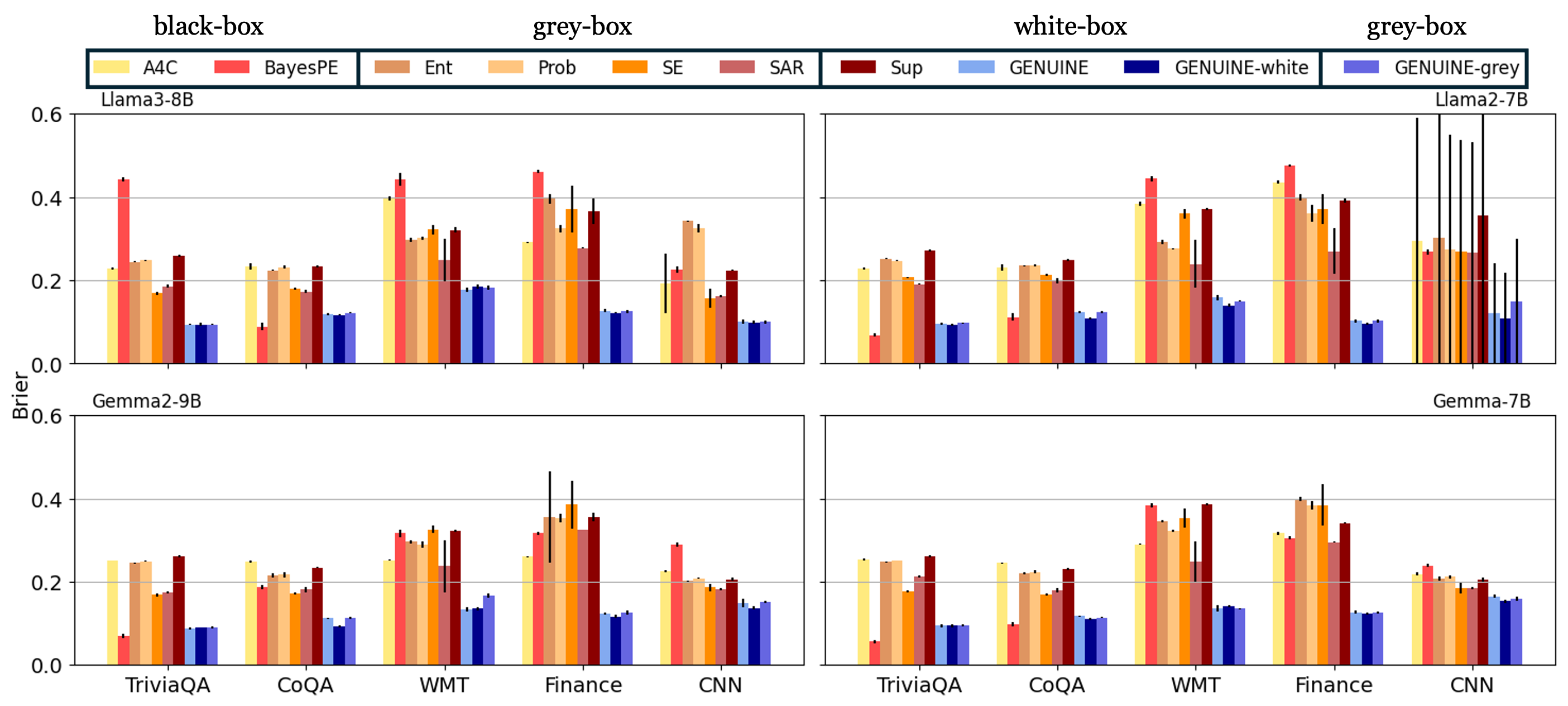}
    \caption{Comparison of Brier scores on five datasets, four LLMs, and seven baselines. Error bars denote variance over five runs.}
    \label{fig: brier results}
\end{figure*}

In this section, we first assess model calibration performance through ECE, NLL, and Brier score metrics, shown in Fig.~\ref{fig: ece results}, Fig.~\ref{fig: nll results}, and Fig.~\ref{fig: brier results}, respectively, comparing \method's reliability against baselines. Then, we present additional experimental results evaluating \method across four key dimensions: (1) the impact of dependency parse trees on uncertainty estimation (Section~\ref{sec: graph structure comparison}), (2) a parameter analysis to determine the sensitivity of \method to hyperparameter tuning (Section~\ref{sec: parameter analysis}), (3) the impact of LLM parameters on \method's uncertainty estimation performance (Section~\ref{sec: parameter impact}), and (4) the impact of training dataset size and noisy labels on \method's uncertainty estimation performance(Section~\ref{ap: section robustness test}).  All experiments are conducted on a Linux server with 64 AMD EPYC 7313 CPUs and an Nvidia Tesla A100 SXM4 GPU with 80 GB of memory.
\begin{figure*}[ht]
    \centering
    \includegraphics[width=\linewidth]{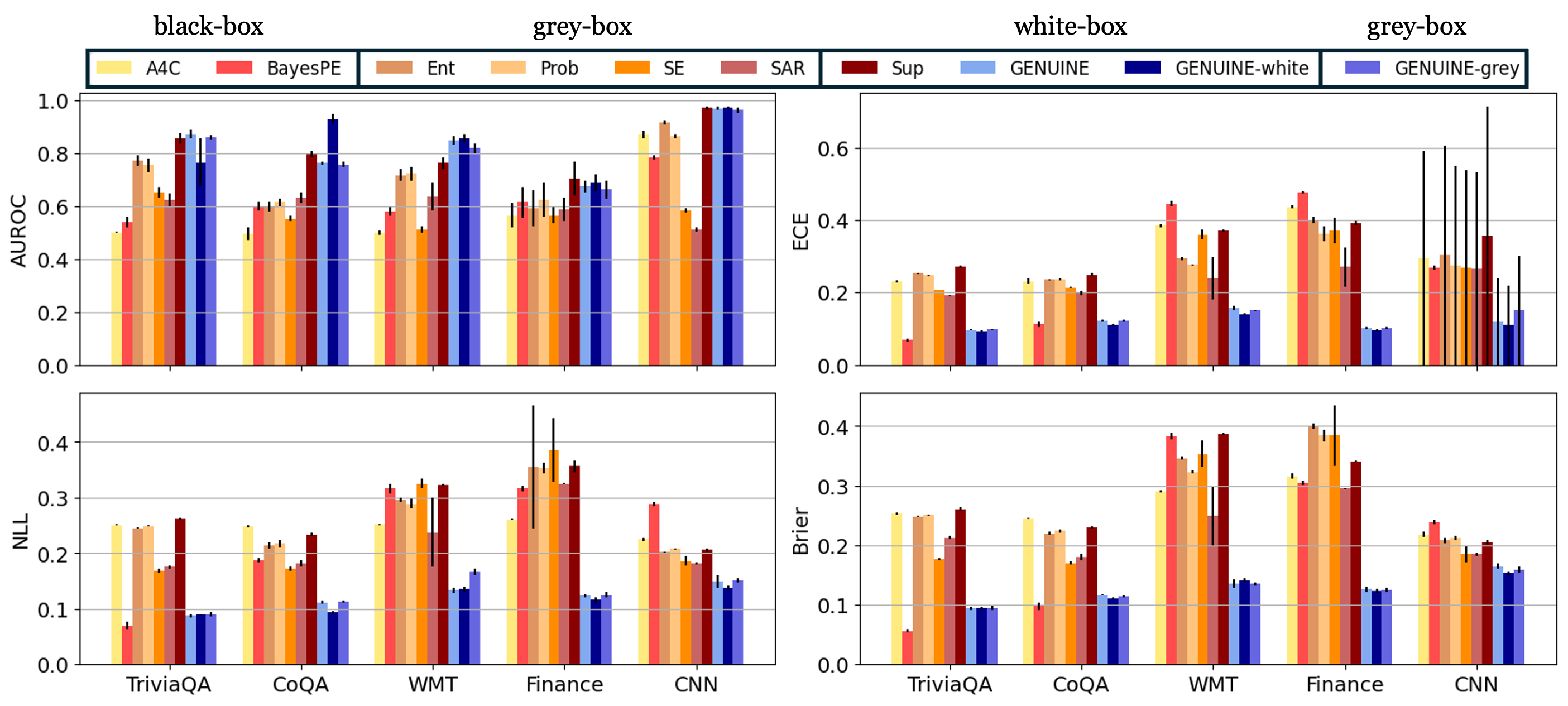}
    \caption{Experimental results on five datasets and seven baseline models on Llama2-13B model. Error bars denote variance over five runs.}
    \label{fig: experimental results on llama2-13B}
\end{figure*}

\begin{table*}[ht]
    \small
    \centering
    \setlength{\tabcolsep}{2pt}
    \caption{Comparison of different graph structures for uncertainty estimation on TriviaQA. NTG refers to the next-token graph utilizing both white-box and grey-box features, while DPT represents the dependency parse tree graph with the same feature set. NTG w/ grey and DPT w/ grey denote the respective graphs using only grey-box features, whereas NTG w/ white and DPT w/ white correspond to configurations using only white-box features.($\uparrow$ means the higher the better, $\downarrow$ means the lower the better)}
    \begin{tabular}{c|c|c|c|c|c|c|c|c}
        \hline
        \multirow{2}{*}{Graphs} & \multicolumn{4}{c|}{Llama3-8B} & \multicolumn{4}{c}{Gemma2-9B} \\
        \cline{2-9}
        ~ & AUROC $\uparrow$ & ECE $\downarrow$ & NLL $\downarrow$ & Brier $\downarrow$ & AUROC $\uparrow$ & ECE $\downarrow$ & NLL $\downarrow$ & Brier $\downarrow$ \\ \hline

        NTG & 0.885±0.048 & 0.264±0.040 & 0.437±0.133 & 0.130±0.062 & 0.846±0.088 & 0.312±0.082 & 0.442±0.122 & 0.131±0.056 \\
        DPT & \textcolor{darkpastelgreen}{0.894±0.032} & \textcolor{darkpastelgreen}{0.246±0.007} & \textcolor{darkpastelgreen}{0.362±0.005} & \textcolor{darkpastelgreen}{0.094±0.002} &  \textcolor{darkpastelgreen}{0.905±0.041} & \textcolor{darkpastelgreen}{0.248±0.009} & \textcolor{darkpastelgreen}{0.356±0.004} & \textcolor{darkpastelgreen}{0.092±0.002} \\
        \hline

        NTG w/ grey & 0.897±0.039 & 0.245±0.007 & 0.363±0.007 & 0.095±0.003 & 0.914±0.041 & 0.251±0.006 & 0.354±0.006 & 0.091±0.003 \\
        DPT w/ grey & \textcolor{darkpastelgreen}{0.903±0.025} & \textcolor{darkpastelgreen}{0.244±0.008} & \textcolor{darkpastelgreen}{0.360±0.003} & \textcolor{darkpastelgreen}{0.094±0.002} & \textcolor{darkpastelgreen}{0.922±0.021} & \textcolor{darkpastelgreen}{0.245±0.005} & \textcolor{darkpastelgreen}{0.352±0.006} & \textcolor{darkpastelgreen}{0.090±0.003} \\ \hline

        NTG w/ white & 0.795±0.049 & 0.249±0.010 & 0.364±0.007 & 0.095±0.003 & 0.960±0.019 & 0.261±0.009 & 0.357±0.006 & 0.092±0.003 \\
        DPT w/ white & \textcolor{darkpastelgreen}{0.809±0.044} & \textcolor{darkpastelgreen}{0.246±0.009} & \textcolor{darkpastelgreen}{0.362±0.007} & \textcolor{darkpastelgreen}{0.094±0.003} & \textcolor{darkpastelgreen}{0.970±0.010} & \textcolor{darkpastelgreen}{0.261±0.006} & \textcolor{darkpastelgreen}{0.353±0.003} & \textcolor{darkpastelgreen}{0.090±0.001} \\ \hline
    \end{tabular}
    \label{tab:graph comparison}
\end{table*}





\begin{figure}
    \centering
    \includegraphics[width=0.9\linewidth]{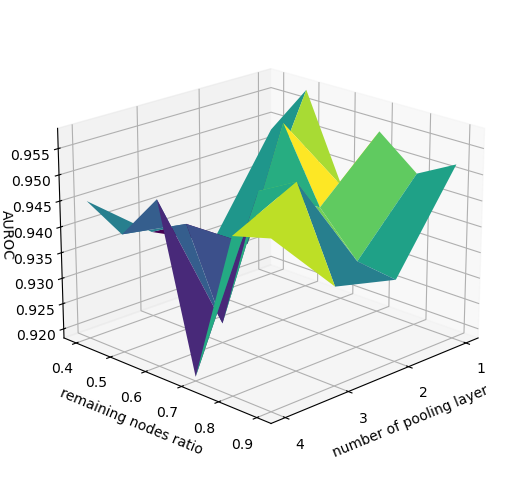}
    \caption{Parameter analysis test on number of pooling layers and remaining nodes ratio for each pooling layer}
    \label{fig: parameter analysis}
\end{figure}

\begin{table}[ht!]
    \small
    \centering
    \caption{Graph Statistics. Here \# Node denotes the average node number and Density denotes the average edge density.}
    \begin{tabular}{c|c|c|c|c}
    \hline
        \multirow{2}{*}{Datasets} & 
        \multicolumn{2}{c|}{Llama3-8B} & \multicolumn{2}{c}{Llama2-7B} \\
        \cline{2-5}
        ~ & \# Node & Density & \# Node & Density \\ \hline

        TriviaQA & 3.86 & 0.56 & 3.77 & 0.58 \\
        CoQA & 5.60 & 0.47 & 5.59 & 0.50 \\
        WMT & 24.01 & 0.11 & 21.75 & 0.13 \\
        Finance & 46.46 & 0.05 & 21.70 & 0.15 \\
        CNN & 61.21 & 0.04 & 87.98 & 0.11 \\
        
        \hline
        \hline

        \multirow{2}{*}{Datasets} & \multicolumn{2}{c|}{Gemma2-9B} & \multicolumn{2}{c}{Gemma-7B} \\
        \cline{2-5}
        ~ & \# Node & Density & \# Node & Density \\ \hline
        TriviaQA & 3.83 & 0.56 & 3.84 & 0.56 \\
        CoQA & 5.28 & 0.47 & 5.19 & 0.48 \\
        WMT & 23.65 & 0.12 & 27.14 & 0.10 \\
        Finance & 43.61 & 0.05 & 42.92 & 0.06 \\
        CNN & 175.33 & 0.01 & 162.96 & 0.01 \\
        \hline
    \end{tabular}
    \label{tab: graphs statistics}
\end{table}

\subsection{Calibration Performance of \method}
\label{calibration_performance}
Calibration ensures that model confidence aligns with actual correctness, making uncertainty estimation more reliable and interpretable. We assess \method and baseline methods using Expected Calibration Error (ECE), Negative Log-Likelihood (NLL), and Brier score, as shown in Fig.~\ref{fig: ece results}, Fig.~\ref{fig: nll results}, and Fig.~\ref{fig: brier results}.

The ECE results (Fig.~\ref{fig: ece results}) reveal that while \method outperforms baselines in WMT, Finance, and CNN datasets, it does not consistently achieve the lowest calibration error in TriviaQA and CoQA. This suggests that token-level methods such as SAR and entropy-based approaches remain competitive in capturing uncertainty effectively for shorter responses. A simple guess that GENUINE underperforms in ECE on TriviaQA is due to the data distribution of the TriviaQA dataset, as shown in Table~\ref{ap: TriviaQA rouge score distribution}. The rouge score for the TriviaQA dataset is not smooth enough, which can bring bias when using the ECE metric. The ECE metric measures the performance based on each bin(group). The number of samples in each bin can be imbalanced due to the distribution of the Rouge score. Thus, we introduce two other calibration metrics, NLL and Brier score, which focus on measuring the calibration gap at the individual level. However, in longer text generation tasks, where error accumulation can distort confidence estimates, \method demonstrates superior calibration by leveraging dependency structures to refine uncertainty aggregation.
\begin{table}[ht!]
    \centering
    \small
    \caption{\textcolor{mycolor}{Distribution of ROUGE score in TriviaQA. The Bin ID indicates the bin index, the Bin Start indicates the start ROUGE score of the selected bin index, and the Bin End indicates the end ROUGE score of the selected bin index. The Density indicates the ratio of samples in this bin index over the total number of samples.}}
    \begin{tabular}{c|c|c|c}
    \hline
        Bin ID& Bin Start & Bin End & Density \\ \hline
        0 & 0.125 & 0.213 & 0.034 \\
        1 & 0.213 & 0.300 & 0.383 \\
        2 & 0.300 & 0.388 & 0.634 \\
        3 & 0.388 & 0.475 & 0.446 \\
        4 & 0.475 & 0.563 & 1.131 \\
        5 & 0.563 & 0.650 & 0.091 \\
        6 & 0.650 & 0.738 & 0.291 \\
        7 & 0.738 & 0.825 & 0.520 \\
        8 & 0.825 & 0.913 & 0.320 \\
        9 & 0.913 & 1.000 & 7.577 \\ \hline
    \end{tabular}
    \label{ap: TriviaQA rouge score distribution}
\end{table}
The NLL results (Fig.~\ref{fig: nll results}) further reinforce these trends. \method consistently achieves lower NLL across all datasets, indicating that it assigns more accurate probability distributions to correct and incorrect responses compared to baselines. The advantage is particularly pronounced in WMT, Finance, and CNN datasets, where long-form responses make token-level uncertainty estimation less effective. Baselines like A4C and SE, which rely on self-evaluation or direct entropy measures, exhibit significantly higher NLL, suggesting that they struggle to generalize confidence estimates across diverse text lengths and response structures.

The Brier score results (Fig.~\ref{fig: brier results}) show that \method achieves competitive performance across all datasets, with particularly strong improvements in WMT, Finance, and CNN datasets, aligning with its NLL performance. The gap between \method and its grey-box and white-box variants indicates that hidden layer representations significantly improve calibration, especially for longer outputs. However, the higher ECE in TriviaQA and CoQA suggests that while structural modeling improves overall uncertainty estimation, it may not always provide the best confidence calibration for shorter text generations, where simpler token-wise approaches remain effective.

These results highlight that \method excels in modeling uncertainty for long-form text but is less dominant in short-response tasks, where entropy-based methods can still provide competitive calibration. The findings reinforce the need for task-specific uncertainty estimation strategies, where dependency-aware modeling is particularly beneficial for applications involving complex text structures and extended reasoning.

\subsection{Graph Structure and Uncertainty Estimation} \label{sec: graph structure comparison ap}
\noindent \textbf{Graph Variations Across Datasets and LLMs.}
Beyond structural differences, graph complexity varies significantly across datasets and LLM architectures, as shown in Table~\ref{tab: graphs statistics}. We observe several key trends.

First, dataset complexity impacts graph structure. TriviaQA produces the shortest outputs, leading to small graphs with an average of 3.8 nodes, while CNN generates significantly longer responses, resulting in much larger graphs (61.2 nodes for Llama3-8B, 175.3 for Gemma2-9B). This confirms that longer text generations create more intricate dependency structures, further reinforcing why graph-based uncertainty estimation is particularly beneficial for longer responses.

Second, LLM architectures influence graph statistics. While Llama models tend to produce slightly longer responses than Gemma models in shorter datasets like TriviaQA and CoQA, this trend reverses in long-form datasets such as CNN, where Gemma models generate significantly longer outputs than Llama models (e.g., 175.3 nodes vs. 61.2 nodes in CNN for Gemma2-9B and Llama3-8B, respectively). This suggests that some LLM families prioritize brevity while others favor more detailed responses, impacting uncertainty estimation requirements.

Lastly, graph density plays a role in structural complexity. Datasets with shorter outputs (TriviaQA, CoQA) tend to have higher edge density, while longer outputs (CNN, Finance) exhibit lower density, indicating that dependency structures become more sparse as response length increases. This suggests that uncertainty estimation models should be designed to handle both dense, local dependencies and sparse, long-range relationships effectively.

\textit{Impact on Uncertainty Estimation Performance:}
The trends in graph statistics correlate directly with AUROC improvements in Fig.~\ref{fig: experimental results}, showing that graph-based uncertainty estimation is particularly beneficial for longer text. The WMT dataset, for example, shows substantial AUROC gains when using graph structures, emphasizing that graph-based methods provide the most value in tasks requiring extended reasoning and structured generation.

Overall, these findings confirm that dependency parsing enhances uncertainty estimation by providing hierarchical token relationships, making it particularly valuable for long-form generation, structured prediction, and document-level tasks. The graph structure directly influences uncertainty estimation effectiveness, reinforcing the need for adaptive modeling strategies based on dataset and model characteristics.
In this section, we want to prove the effectiveness of the dependency parse tree and show the different patterns in graph structures from different datasets and different LLMs as well.

\subsection{Ablation Study} \label{ap: alabtion study}
\textcolor{mycolor}{To further prove the effectiveness of the graph structure and the fused assignment matrix, we offer more ablation experiments on the TriviaQA dataset using Llama2-7B and Gemma-7B. As shown in Table~\ref{tab: ablation on fusion process in ap}, the fusion process (Fig.~\ref{fig:framework}) improves AUROC by 2.02\% for Llama2-7B and 1.89\% for Gemma-7B, the graph structure process improves AUROC by 1.0\% for Llama2-7B and 5.5\% for Gemma-7B.
These results demonstrate that the graph structure and the fusion strategy effectively integrate structural and semantic uncertainty signals, enabling more robust uncertainty propagation across tokens. In contrast, methods w/o a graph structure and fusion strategy fail to capture meaningful relationships between uncertainty features, leading to suboptimal performance.
The consistent improvement across models highlights the importance of structured features and the fusion process in uncertainty estimation. By jointly optimizing structural and semantic representations, \method enhances both robustness and interpretability, making it well-suited for uncertainty-aware applications.}

\begin{table}[t!]
    \centering
    \small
    \caption{Ablation study of fusion process on TriviaQA($\uparrow$ means the higher the better)}
    \begin{tabular}{c|c|c}
    \hline
        \multicolumn{3}{c}{TriviaQA} \\ \hline
        \multirow{2}{*}{Methods} & Llama3-8B & Llama2-7B \\
        \cline{2-3}
        ~ & AUROC $\uparrow$ & AUROC $\uparrow$ \\ \hline
        \makecell{\method w/o \\ fusion \& graph} & 0.789±0.031 & 0.835±0.005 \\
        \method w/o fusion & 0.809±0.096 & 0.843±0.011 \\ 
        \method & \textcolor{darkpastelgreen}{0.894±0.032} & \textcolor{darkpastelgreen}{0.860±0.027} \\
        \hline
        \hline
        \multirow{2}{*}{Methods} & Gemma2-9B & Gemma-7B \\
        \cline{2-3}
        ~ & AUROC $\uparrow$ & AUROC $\uparrow$ \\ \hline
        \makecell{\method w/o \\ fusion \& graph} & 0.956±0.002 & 0.853±0.033 \\
        \method w/o fusion & 0.963±0.015 & 0.900±0.037 \\ 
        \method & \textcolor{darkpastelgreen}{0.969±0.009} & \textcolor{darkpastelgreen}{0.917±0.047} \\
        \hline
        \hline
        \multicolumn{3}{c}{WMT} \\ \hline
        \multirow{2}{*}{Methods} & Llama3-8B & Llama2-7B \\
        \cline{2-3}
        ~ & AUROC $\uparrow$ & AUROC $\uparrow$ \\ \hline
        \makecell{\method w/o \\ fusion \& graph} & 0.709±0.013 & 0.844±0.020 \\
        \method w/o fusion & 0.713±0.002 & 0.850±0.004 \\ 
        \method & \textcolor{darkpastelgreen}{0.826±0.019} & \textcolor{darkpastelgreen}{0.853±0.015} \\
        \hline
        \hline
        \multirow{2}{*}{Methods} & Gemma2-9B & Gemma-7B \\
        \cline{2-3}
        ~ & AUROC $\uparrow$ & AUROC $\uparrow$ \\ \hline
        \makecell{\method w/o \\ fusion \& graph} & 0.898±0.012 & 0.736±0.033 \\
        \method w/o fusion & 0.905±0.002 & 0.743±0.007 \\ 
        \method & \textcolor{darkpastelgreen}{0.914±0.014} & \textcolor{darkpastelgreen}{0.837±0.023} \\
        \hline
    \end{tabular}
    \label{tab: ablation on fusion process in ap}
\end{table}

\subsection{Parameter Sensitivity} \label{sec: parameter analysis}
Understanding the impact of hyperparameters on \method's performance is essential for optimizing uncertainty estimation while ensuring efficiency. We evaluate two key parameters: the number of pooling layers (ranging from 1 to 4) and the remaining node ratio at each pooling step. The results, shown in Fig.~\ref{fig: parameter analysis}, reveal important trends that highlight \method's robustness and adaptability.

The results indicate that AUROC remains high with fewer pooling layers, suggesting that a deep hierarchy is not necessary for effective uncertainty estimation. As the number of pooling layers increases, performance fluctuates, indicating that excessive pooling may lead to loss of critical structural information, reducing the model's ability to capture meaningful uncertainty signals. This trend suggests that \method achieves optimal results with a moderate number of pooling layers, avoiding unnecessary complexity while maintaining strong predictive performance.

Additionally, the remaining node ratio plays a crucial role in uncertainty estimation. The model may struggle with redundant information when too many nodes are retained, leading to slightly lower AUROC. However, when the number of retained nodes is optimized, performance improves, reinforcing the idea that removing less informative nodes enhances uncertainty representation. Interestingly, when the remaining ratio is lower, but the number of pooling layers is set appropriately, AUROC reaches peak performance, highlighting the benefits of structured feature reduction in refining uncertainty quantification.

Overall, these findings demonstrate that \method is robust to hyperparameter choices, requiring minimal tuning to achieve strong performance. The ability to maintain high AUROC across a range of configurations suggests that \method can be easily applied to various tasks and LLMs without extensive parameter optimization, making it highly adaptable for real-world deployment.

\subsection{Impact of LLM Parameters} \label{sec: parameter impact}

Understanding how LLM architecture and scale affect uncertainty estimation is crucial for assessing the generalizability of \method. We compare the performance of Llama2-13B (Fig.~\ref{fig: experimental results on llama2-13B}) against Llama3-8B and Llama2-7B, analyzing its effectiveness across AUROC, calibration metrics (ECE, NLL, and Brier scores), and overall robustness.

\noindent \textbf{Uncertainty Estimation Across LLM Variants.}  
Llama2-13B achieves strong AUROC performance across all datasets, often matching or surpassing Llama3-8B and Llama2-7B. The improvements are particularly evident in WMT, Finance, and CNN datasets, where Llama2-13B consistently outperforms its smaller counterparts. This suggests that larger models benefit from enhanced representation learning, leading to more stable and accurate uncertainty estimation in complex, long-form text generation tasks. However, in TriviaQA and CoQA, the AUROC gains are marginal, indicating that the advantages of increased model size are less pronounced for shorter responses.

\noindent \textbf{Calibration Trends: ECE, NLL, and Brier Score Analysis.}  
One notable observation is that \method outperforms baselines in ECE for TriviaQA and CoQA on Llama2-13B, whereas this trend is not observed in Llama3-8B and Llama2-7B. This suggests that larger models may allow \method to better align confidence scores with correctness probabilities in short-response tasks, where previous versions struggled to outperform entropy-based baselines. The ECE results (Fig.~\ref{fig: ece results}) further confirm that in WMT, Finance, and CNN, Llama2-13B achieves lower calibration errors, highlighting its ability to generate better-aligned confidence estimates for longer outputs.

The NLL and Brier score results (Fig.~\ref{fig: nll results} and Fig.~\ref{fig: brier results}) reinforce these findings. Llama2-13B consistently achieves lower NLL and Brier scores across datasets, particularly in WMT, Finance, and CNN, where uncertainty estimation benefits from structured confidence propagation. This suggests that larger models improve AUROC and provide better-calibrated uncertainty estimates, making them well-suited for tasks requiring complex reasoning and structured text.

The results indicate that larger models significantly enhance both uncertainty estimation and confidence calibration, particularly in short-response tasks like TriviaQA and CoQA, where \method surpasses entropy-based baselines in ECE for the first time. This suggests that model size can influence calibration effectiveness differently across datasets, with larger architectures improving both long-form uncertainty quantification and short-text confidence alignment. Future research should explore adaptive calibration strategies tailored to different response lengths, ensuring that LLMs remain reliable across diverse NLP applications.

Overall, these findings reinforce that \method scales effectively across different LLM architectures, maintaining robust uncertainty estimation and calibration performance while highlighting areas where model size influences uncertainty quantification.
\begin{table}[h!]
    \centering
    \small
    \caption{\textcolor{mycolor}{The impact of training dataset size on \method performance. More training data results in higher AUROC, but no significant decrease of AUROC when using at least 20\% of the training data. Training dataset size does not have much influence on the calibration metrics($\uparrow$ means the higher the better, $\downarrow$ means the lower the better)}}
    \begin{tabular}{c|c|c}
    \hline
        training size & AUROC$\uparrow$ & ECE$\downarrow$ \\ \hline
        0.1 & 0.813±0.059 & 0.239±0.009 \\
        0.2 & 0.873±0.020 & 0.244±0.007 \\
        0.3 & 0.883±0.012 & 0.241±0.011 \\
        0.4 & 0.854±0.056 & 0.241±0.007 \\
        0.5 & 0.874±0.015 & 0.243±0.010 \\
        0.6 & 0.894±0.032 & 0.246±0.007 \\ \hline
        \hline
        training size & NLL$\downarrow$ & Brier$\downarrow$ \\ \hline
        0.1 & 0.361±0.006 & 0.094±0.003 \\
        0.2 & 0.363±0.006 & 0.095±0.003 \\
        0.3 & 0.363±0.001 & 0.095±0.000 \\
        0.4 & 0.360±0.005 & 0.094±0.002 \\
        0.5 & 0.362±0.004 & 0.094±0.002 \\
        0.6 & 0.362±0.005 & 0.094±0.002 \\ \hline
    \end{tabular}
    \label{ap: training size impact on performance}
\end{table}

\subsection{Robustness Test on Training Dataset Size and Noisy Labels} \label{ap: section robustness test}
\textcolor{mycolor}{In real-world scenarios, uncertainty estimation models often face limited training data and noisy labels, which can affect performance. To evaluate the robustness of \method under such conditions, we conduct experiments using the Llama3-8B model on the TriviaQA dataset. Table~\ref{ap: training size impact on performance} shows how varying training set sizes impact performance, while Table~\ref{ap: noisy label impact on performance} examines the effect of label noise. For the latter, we randomly corrupt a portion of training labels (as specified by the noise ratio) and assess performance on the clean test set. These experiments demonstrate \method’s resilience to data scarcity and label noise, highlighting its applicability in real-world settings.}

\textcolor{mycolor}{The results shown in Table~\ref{ap: training size impact on performance} indicate that the number of training samples does influence \method's performance, especially when using only 10\% of the training data, the AUROC drops 9.1\% compared to the model using 60\% training data. However, when the training samples take between 20\% and 50\% of the whole samples, the performances remain relatively stable. Another observation is that the training dataset size does not have much influence on the ECE, NLL, and Brier score.}

\textcolor{mycolor}{From the results in Table~\ref{ap: noisy label impact on performance}, we find that the noisy labels have a negative influence on the models' performance in general. However, \method remains robust when 0.1\% of the training samples are polluted. As the noise ratio increases, the AUROC drops significantly, as well as the ECE, NLL, and Brier score. We can conclude that, unlike training dataset size, which has little impact on the calibration metrics, the noise ratio influences not only the AUROC but also the calibration results.}

\end{document}